# Atrial Fibrillation: A Medical and Technological Review

Samayan Bhattacharya
Department of Computer Science and Engineering
Jadavpur University
188, Raja Subodh Chandra Mallick Rd, Jadavpur University Campus Area, Jadavpur, Kolkata, West Bengal 700032, India
samayan.bhattacharya@gmail.com

Sk Shahnawaz
Department of Computer Science and Engineering
Jadavpur University
188, Raja Subodh Chandra Mallick Rd, Jadavpur University Campus Area, Jadavpur, Kolkata, West Bengal 700032, India
skshahnawaz2909@gmail.com

## ABSTRACT

Atrial Fibrillation (AF) is the most common type of arrhythmia (Greek a-, loss + rhythmos, rhythm = loss of rhythm) leading to hospitalization in the United States. Though sometimes AF is asymptomatic, it increases the risk of stroke and heart failure in patients, in addition to lowering the health-related quality of life (HRQOL). AF-related care costs the healthcare system between $6.0 to $26 billion each year[1]. Early detection of AF and clinical attention can help improve symptoms and HRQOL of the patient, as well as bring down the cost of care. However, the prevalent paradigm of AF detection depends on electrocardiogram (ECG) recorded at a single point in time and does not shed light on the relation of the symptoms with heart rhythm or AF. In the recent decade, due to the democratization of health monitors [10] and the advent of high-performing computers, Machine Learning algorithms have been proven effective in identifying AF, from the ECG of patients. This paper provides an overview of the symptoms of AF, its diagnosis, and future prospects for research in the field.

## 1. Introduction

Atrial Fibrillation is an irregular and elevated heart rate in which the upper chambers of the heart (atria) beat out of coordination with the lower chambers of the heart (ventricles). Symptoms typically include shortness of breath, heart palpitation, and weakness. Though AF by itself is not life-threatening, it increases the risk of stroke, heart failure, and other complications. Episodes of AF may come and go in some patients, while in others, it may persist and require medical intervention. A major concern of AF is the formation of clots in the atria. These clots may travel to other organs and block off blood flow. AF is treated with medication and other interventions to restore normal electrical activity in the heart.

## 1.1 Mechanism of Atrial Fibrillation

In a person without AF, the contractions of the different chambers of the heart are coordinated by the heart's electrical system. The impulses originate in the sinoatrial (SA) node, a group of special cells in the right upper chamber (atrium) of the heart, in an orderly fashion and at a rate depending on the activity the person is performing (rate is higher while running compared to that while sleeping). The impulses then travel through a network of conducting cell "pathways" and causes the atria to contract and squeeze blood into the ventricles. The impulses then travel to the ventricles through the atrioventricular (AV) node. The impulses then spread across the ventricles through the His-Purkinje Network. This causes the ventricles to contract and squeeze blood out to the lungs and rest of the body. The normal heart beats about 60 to 100 times per minute at rest.

In patients with AF, the SA node produces impulses in a rapid and chaotic manner. The atria contract with a chaotic rhythm and are unable to effectively squeeze blood into the ventricles. Instead of the impulses flowing to the ventricles in an orderly fashion, several impulses compete to enter the ventricles through the AV node. The AV node acts as a filter and allows only some of the impulses to pass. These cause the ventricles to contract irregularly and out of synchronization with the atria. The rate of impulses can range from 300 to 600 beats per minute with the atria and ventricles beat at different rates.

## 2. Types of Atrial Fibrillation

The American heart association recommends the classification of AF into 4 categories: first detection, paroxysmal, persistent, and permanent, based on the temporal rhythm due to its clinical relevance [2].

### 2.1 First Detection

The first time AF is detected in a person, irrespective of the duration of the episode is called First Detection AF. It may be symptomatic or asymptomatic. The classical way of diagnosing AF based on the ECG of the patient is not preferred as it is relevant only when serious symptoms are seen.

- 2.2 Paroxysmal

    Paroxysmal AF that stops without any intervention within a week. This type of AF appears repeatedly if left untreated and hence early detection is critical.

- 2.3 Persistent

    This type of AF persists for longer than a week and recurrence is possible.

- 2.4 Permanent

    If AF symptoms do not get better even with medical intervention or the symptoms get better only temporarily, it is said to be Permanent AF.

# 3. Symptoms

Due to the dominant paradigm of diagnosing AF based on the ECG recorded at a single point in time instead of in real-time corresponding to the observation of the symptoms , there is a paucity of data relating to particular symptoms with AF. The common symptoms associated with AF are alterations in sympathetic nervous system function, impaired myocardial perfusion, decreased cardiac output, and impaired ventricular diastolic filling.[4,5,6,7]. The classical method of judging the effectiveness of AF treatment procedures like cardioversion ablation and Cox-Maze procedure is by observing the occurrence of symptoms in the time between clinic visits or by getting an ECG or limited Holter monitor recordings of the patient at their next clinic visit to get a clearer picture of heart rhythm after the procedure. However, these procedures are unable to shed light on asymptomatic episodes of AF, which continues to pose the threat of stroke, pulmonary embolism, and deep vein thrombosis. Studies with wearable heart rate monitoring devices (Eg. smartwatches) and implantable cardiac monitoring systems have shed light on the true recurrence of AF. In a study Page et al found that in patients with symptomatic paroxysmal AF monitored for 12 months, asymptomatic atrial tachyarrhythmia recurred 12 times more often than symptomatic tachyarrhythmia [9]. Later studies found that in patients with other symptomatic diseases, asymptomatic AF occurred far more often, accounting for 54-94% of all AF arrhythmias.[11,12,13,14] Asymptomatic AF is more prevalent among patients who have undergone some interventional clinical procedure for the treatment of AF. . According to Verma and colleagues, the ratio of asymptomatic to symptomatic occurrences of AF increased from 1.1 to 3.7 following catheter ablation of the patient [7].

The use of implantable cardiac monitoring devices has not only allowed us to pick up on asymptomatic AF but also allowed us to try to correlate the observed symptoms with heart rhythms. Two separate studies attempting to find the predictive value of reported symptoms to heart rhythm reported a predictive value of only 17-21% [11,14]. Results obtained from varied paroxysmal AF patients showed that of all symptom reports, 45-79% were reported when there was no device-confirmed AF episode[11,12,14]. Causal factors of non-AF arrhythmic symptoms have yet to be explained. Regardless of actual heart rhythm during symptom presentation, the goal of treatment of AF remains measurement and management of symptoms.

## 4. Clinical diagnosis

Presently, the detection of AF is performed based on the electrocardiogram (ECG) of the patient. ECG shows the variation of voltage with respect to time and hence reflects the depolarization followed by repolarization of the heart muscles at the time of each heartbeat. The ECG graph of a normal beat (shown in Fig. 1) consists of a sequence of waves, a V-wave presenting the atrial depolarization process, a WXY complex denoting the ventricular depolarization process, and a Z-wave representing the ventricular repolarization. Other portions of the signal include the VX, YZ, and WZ intervals.

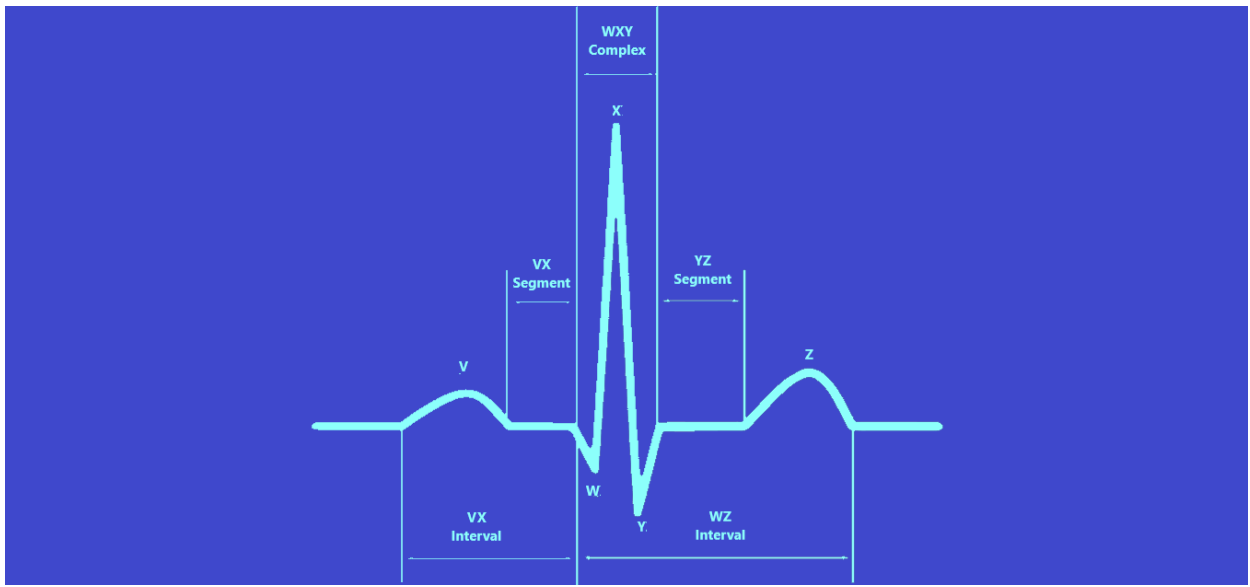

Figure 1: ECG of normal heartbeat

## 5. Role of Machine Learning

Machine Learning is a subfield of Artificial Intelligence and it allows computers to learn from data. Deep Learning is a subfield of Machine Learning that uses multiple layers of computational units, called neurons. Training a deep learning model involves showing it lots of examples of the kind of data it is to learn and then it figures out a relationship between the inputs and the outputs. The deep learning model starts with random weights for the function it is trying to learn and then as it sees an example, it takes the input and calculates the output. Then it measures how far off the calculated output is from the expected output using a loss function and then adjusts the weights using backpropagation of the loss through the layers of neurons to reduce the loss [25].

In the early days of the application of Machine learning, the machine learning models used to rely on humans to select the features the model is to learn from. For example, in ECG analysis, the model would be provided with the inputs as morphological and temporal features and the outputs to be predicted would be the ECG rhythm, the serum potassium level, or the LV ejection fraction (LVEF). The disadvantage of this method was that humans were dependent on their experience and the available knowledge to select features, which were both unreliable and varied among individuals.

Thus, in deep learning, the features are selected by the model itself during training. The most common model for AI-ECG is called a Convolutional Neural Network (CNN). It uses convolutions, in the form of filters, to select the features to be considered while predicting the output. The weights of the convolutional layers adjust the contribution of each part of the input matrix (image, ECG, or any data that can be represented in 2D) towards predicting the output. The neural network can be thought of as a sequence of extraction layers (the convolutional layers) and mathematical layers (like pooling layers, rectified linear unit, dropout, etc.), which accept the features selected by an extraction layer and perform the corresponding operations to give an output. The organization of these layers, the number, and the shape of the convolutional layers are selected by the creator of the network based on intuition and trial-and-error outcomes[26].

The ECG data represents the time coordinate along the horizontal axis and the recorded voltage along the vertical axis. Thus, convolution can be vertical, accepting the voltage values from all the leads, or horizontal, accepting the voltage values from a single lead across all time points or it may be 2D to accept both time and voltage information of all leads across all time points.

Thus deep learning allows the network to select features and learn from them without human bias and limitations stemming from the lack of certain information to humans. This agnostic approach to solving the problem is an optimal one, presently, but it conceals the inner workings of the neural network. Humans are unable to figure out why the network is looking at a piece of data and making a particular prediction. This operation as a black box raises concerns in the medical community[15]. Thus, less agnostic models like the random forest, reinforcement learning, and logistic regression still appear promising to researchers and clinical practitioners.

Reinforcement learning is an area of machine learning where the model is rewarded in proportion to the correctness of the predicted output[16]. The model learns to take the steps necessary to maximize the reward. Random forest involves the construction of many decision trees during execution[17]. The final output is the output predicted by the highest number of trees.

Another disadvantage of the CNN architecture is that it is a supervised learning algorithm. Unlike, semi-supervised and unsupervised learning algorithms, all the training data for a CNN model needs to be labeled. This is time-consuming. Unsupervised learning algorithms, like clustering algorithms, learn to figure out common attributes of inputs belonging to the same class[18]. Thus, unlabelled data can be used to train the model.

In addition to the types of neural networks discussed above, additional input might also be included, like natural language processing might be used to process text, including prescriptions, medical records, and description of symptoms[19]. These methods include rule-based recognition of topic modeling, text vectorization, and word patterns. Then an ensemble of the output of the various networks can be used to predict the final output.

## 5.1     Wearable devices and mobile ECG technologies

AI algorithms can be applied to wearable and portable devices, allowing point-of-care diagnosis for patients. Though most algorithms are designed for 12-lead ECG data, algorithms are deployed on single-lead [22]. In addition to ECG data, other data can also be analyzed with these algorithms. For example, Apple Watch uses photoplethysmography signals to detect AF passively[10]. Also, these algorithms have been used to detect hypertrophic cardiomyopathy (HCM) or the determination of serum potassium levels using single-lead ECG [20,21]. Recent versions of the Apple watch allow the user to confirm the presence of AF using electrophysiological signals obtained from a single bipolar vector[10].

## 5.2     Advantages and Challenges

AI models offer higher diagnostic fidelity and workflow efficiency than human diagnosticians. ECG data, due to its simplicity and low memory requirements is ideal for training Machine learning algorithms. The large databanks available to researchers in addition to the high capabilities of modern computers have allowed machine learning algorithms to outperform humans. Machine learning algorithms are able to spot very subtle patterns in the ECG that allow them to detect AF even when ECG is recorded when the patient is not having an AF episode [24]. In addition to AF, these patterns can also be used to detect other heart diseases like hypertrophic cardiomyopathy (HCM), silent atrial fibrillation (AF), and left ventricular (LV) systolic dysfunction, and might also shed light on systemic physiology, such as a person's sex, age or their serum potassium levels.

Wearable health monitoring devices have a size limitation which translates to a memory limitation. However, so long the device is connected with another device having a significant amount of memory (like a smartphone), the recorded data can be used in a multitude of ways. Surveillance of the WV interval for patients in whom treatment with drugs that affect repolarization (Eg. sotalol, dofetilide ) is initiated can help report the reaction of AF symptoms to the drugs in real-time. Patients, with hypertrophic cardiomyopathy or inherited arrhythmia syndromes, at risk of cardiac failure can be monitored for stress levels and early warning can be issued to take precautions. In case of the sudden collapse of a patient, at risk of cardiac failure, real-time data can be accessed immediately for quick medical assistance, without having to wait for ECG results. These algorithms might also be able to predict AF events and prescribe pills.[24]

The machine learning algorithms are dependent on large quantities of data to be able to train and operate to satisfactory standards. The challenges in the wide-scale deployment of these algorithms are mostly the challenges involved in collecting proper data. The data collected from a certain population and used to train the model might produce a model that performs well for that population but poorly for other populations. Also, for a given population, the data might change over time due to environmental factors. There is a chance that the people creating the model might ignore some parameter that allows the model to perform better. Clearly, the advent of AI-related applications in the clinical environment generates a new set of previously unrecognized challenges.

# 6. Future Research Opportunities

There is a wide spectrum of research going on to build better, more efficient AI models. As the world becomes more connected and computers become more accessible to people, researchers are able to access more diverse datasets that help reduce biases in their models. Thus, application of AI in ECG data processing (here on referred to as AI-ECG) would better operate on a wide variety of image qualities as well as ECG of different populations. As with other applications of AI in medicine, the explainability of models would be an important part of getting public trust.

The interdisciplinary nature of AI-ECG presents its own challenges. To make the transition from traditional diagnosis to AI-ECG, it is important that those who are creating the technology must make it intuitive enough so that it is easy for organizations to get people familiarised with new technology, so as to let them operate it successfully. A major hurdle on the road to large-scale deployment of AI-ECG would be regulatory approvals. Like any other medical procedure, AI-ECG is important to be vetted, validated, and verified before deployment. As we venture into new territories, it is hard to figure out the right questions to be asked for regulatory approval. Close cooperation of industries and governments would be of paramount importance.

## Acknowledgement

We would like to acknowledge Mr.Rangan Das, PhD candidate, CSE, Jadavpur University, for his constant support and motivation.